# Prior-AttUNet: Retinal OCT Fluid Segmentation Based on Normal Anatomical Priors and Attention Gating

Li Yang[a,b], Yuting Liu[a,c,*]

[a]School of Medical Information, Wannan Medical College, Wuhu, 241002, China
[b]Department of Information, Wuhu Shengmeifu Technology Co. Ltd, Wuhu, 241002, China
[c]Department of Information, Anhui Taihao Intelligent Technology Co. Ltd, Wuhu,241002, China
*Corresponding author. E-mail: liuyuting@wnmc.edu.cn

**Abstract**

Macular edema is a key pathological feature of blinding eye diseases such as age-related macular degeneration and diabetic macular edema, and its precise segmentation is crucial for clinical diagnosis and treatment. To address challenges in fluid segmentation within optical coherence tomography (OCT) images, such as blurred boundaries and inter-device variations, this paper proposes a Prior-AttUNet model enhanced by generative priors. The model constructs a hybrid generative-segmentation dual-path architecture, which provides multi-scale normal anatomical priors through a variational autoencoder and embeds densely connected blocks along with spatial pyramid pooling modules in the backbone network to enhance contextual awareness. Furthermore, an anatomical prior-guided triple attention mechanism is designed to adaptively adjust feature weights across multiple decoding stages, significantly improving segmentation accuracy at lesion boundaries. Experimental results on the RETOUCH dataset demonstrate that Prior-AttUNet performs excellently on three imaging devices—Cirrus, Spectralis, and Topcon—achieving average Dice similarity coefficients of 93.93%, 95.18%, and 93.47%, respectively. Moreover, its computational complexity is only 0.37 TFLOPs, achieving an optimal balance between segmentation accuracy and inference efficiency, thereby providing a reliable solution for automated clinical diagnosis.

## 1. Introduction

Macular edema is a common and critical pathological manifestation of blinding eye diseases such as age-related macular degeneration (AMD) and diabetic macular edema (DME). It is characterized by the presence of different types of fluid accumulations in the macular region of the retina, including intraretinal fluid (IRF), subretinal fluid (SRF), and pigment epithelial detachment (PED) [1]. Accurate assessment of these fluid regions is crucial for evaluating visual prognosis and guiding anti-VEGF treatment plans [2]. Optical coherence tomography (OCT), as a high-resolution in vivo tissue imaging technique, has become the gold standard for clinical diagnosis and monitoring of macular edema [3-4].

In clinical practice, precise segmentation of fluid in OCT images faces multiple challenges. Firstly, relying on manual annotations by ophthalmologists is not only time-consuming and labor-intensive but also subject to subjective bias, making it difficult to apply efficiently and reproducibly across large patient populations. Secondly, different types of fluid may exhibit features in OCT images such as low contrast and blurred boundaries, making them inherently difficult to distinguish from each other, posing a fundamental challenge for accurate segmentation [5]. Furthermore, OCT images themselves are susceptible to interference from eye motion artifacts and speckle noise, further compromising analytical accuracy. Finally, differences in imaging parameters (e.g., resolution, pixel distribution) among OCT devices from various manufacturers lead to significant variations in image characteristics [6]. The three images in Figure 1 are sourced from the RETOUCH dataset [7], captured by Cirrus, Spectralis, and Topcon devices from left to right, respectively. The green areas represent IRF, orange areas represent SRF, and yellow areas represent PED. It is evident that the morphological differences between different fluid types are significant, while the pixel distribution characteristics also vary greatly across devices. These factors collectively pose a severe challenge to developing automated tools with broad generalization capabilities.

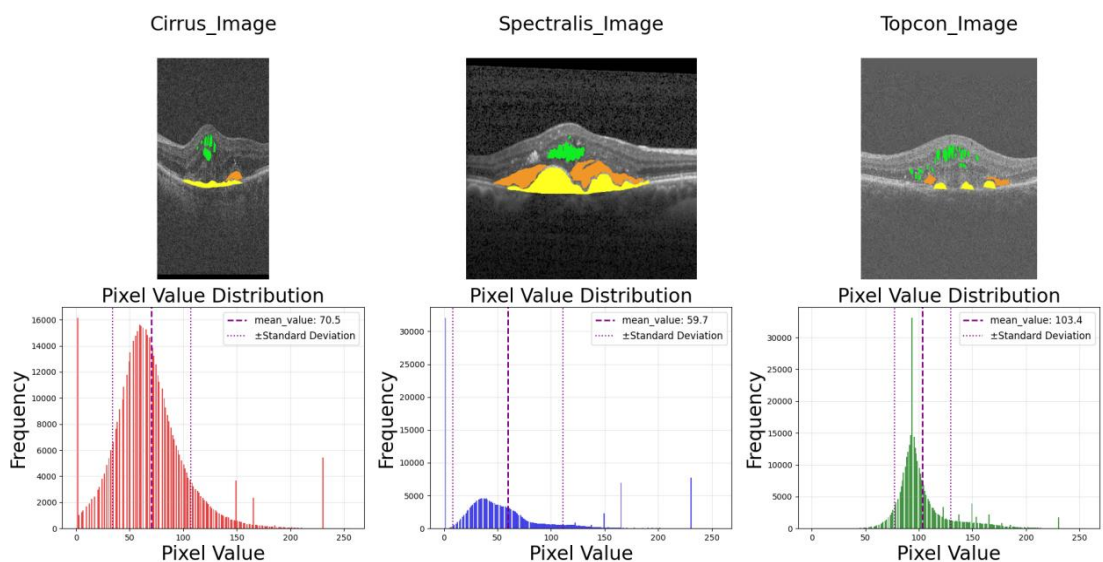

**Figure 1:** Differences in fluid morphology and pixel intensity distribution in OCT images across different imaging devices.

With the deepening application of deep learning in medical image analysis, advanced deep network architectures such as UNet [8], Attention U-Net [9], DeeplabV3+ [10], and nnUNet [11] have been successfully applied to fluid segmentation tasks in OCT images. However, most existing methods [2,5,6,12] primarily focus on enhancing the extraction of fluid features by optimizing the internal network structure, often overlooking valuable external prior information—particularly the normal anatomical representation of healthy retinas. This limitation stands in stark contrast to clinical diagnostic practice: physicians typically require extensive training on normal imaging cases to accurately identify pathological regions. Based on this clinical insight, such contrastive detection offers a novel solution for medical segmentation [13-15].

Kobayashi et al. [13] proposed a discriminative network approach that first reconstructs normal brain image patches in the feature space, then measures the L2 distance between tumor image patches and the reconstructed normal patches, ultimately utilizing this difference metric to achieve block-level localization of brain tumors. Astaraki et al. [14] introduced a method based on a variational autoencoder (VAE) [16]. This method first employs a VAE to generate normal, lesion-free lung images corresponding to the input images. Subsequently, during the segmentation process, systematic comparison between the original tumor lung images and the generated normal images effectively enhances segmentation accuracy for lung tumor regions. Liu et al. [15] proposed an innovative segmentation framework based on IntroVAE [17]. This framework enhances the representation of tumor features by introducing unimodal normal brain images as references in the feature space for comparison with multimodal tumor images, while utilizing a feature alignment module (FAM) to address the discrepancies between them.

The core principle of this study is to utilize normal tissue images reconstructed by generative neural networks [16-18] as a reference baseline. By systematically comparing these with actual medical images containing lesions, precise localization of the pathological regions is achieved.

Based on the core concepts of normal anatomical priors and attention gating, this study proposes a segmentation method for retinal OCT images called Prior-AttUNet. This method uses IntroVAE to reconstruct fluid-free OCT images as a reference baseline. Through contrastive learning between normal OCT features and fluid region features, it effectively enhances the feature representation capability of fluid regions, thereby significantly improving segmentation performance. The model adopts a dual-path generation-segmentation architecture, using U-Net as the basic framework. Lightweight dense connection blocks based on depthwise separable convolutions [19-21] are introduced in the encoder and decoder to enhance feature reuse and propagation efficiency. An ASPP module [9] is embedded in the bottleneck layer to improve multi-scale contextual awareness. Additionally, a novel triple attention gating mechanism is designed to fuse skip connection features, decoder upsampling features, and normal anatomical priors generated by the VAE, enabling finer feature selection and boundary perception.

**The main contributions of this paper are as follows:**

**(a)A generative prior-enhanced multi-scale architecture.** We construct a hybrid generative-segmentation dual-path architecture that integrates a VAE-based generative network with an enhanced U-Net. This architecture leverages the VAE branch to provide multi-scale normal anatomical priors, while embedding densely connected blocks and an ASPP module into the backbone network to collectively enhance the model's contextual awareness and segmentation robustness for pathological structures.

**(b) An anatomically prior-guided triple attention mechanism.** We design a triple attention gating mechanism that, in addition to traditional skip connections and decoder features, introduces normal anatomical priors generated by the VAE as guidance. Through the synergy and suppression relationships among features, this mechanism enables more refined feature selection and adaptively adjusts weights across multiple decoding stages, thereby significantly improving segmentation accuracy at the boundaries of lesion regions.

**(c) Data-driven global optimization based on systematic ablation experiments.** Through ablation experiments covering both component functionality and architectural parameters, we achieve data-driven global optimization: first validating the effectiveness of core components (e.g., triple attention gating, multi-scale aggregation modules), and then finely tuning key parameters such as anatomical prior generation and dense connectivity patterns. This process collectively ensures optimal performance and efficiency of the model.

## 2. Related Work
### 2.1. Fluid Segmentation Methods

In retinal OCT images, intraretinal fluid (IRF) is closely associated with diabetic macular edema and diabetic retinopathy; subretinal fluid (SRF) often serves as an important indicator for postoperative prognosis in age-related macular degeneration; while pigment epithelial detachment (PED) is primarily caused by conditions such as polypoidal choroidal vasculopathy, retinal angiomatous proliferation, and central serous chorioretinopathy [22-26]. Accurate quantification of these fluids holds significant importance in the early diagnosis of related pathologies.

Numerous advanced deep learning methods—such as FCN [27], UNet, Attention U-Net, nnUNet, SegNet [28], Deeplabv3+, and UNet++ [29]—have been applied to fluid segmentation in OCT images.

To enhance model performance, researchers have tailored segmentation networks by incorporating attention mechanisms, feature propagation mechanisms, and context-aware modules. These designs effectively address issues such as blurred boundaries of fluid regions and insufficient contrast between different fluid types, thereby significantly improving the generalization capability of the models.

Among various attention mechanisms, the research community has proposed a range of representative methods, including Squeeze-and-Excitation Networks (SENet) [30], Efficient Channel Attention Network (ECANet) [31], and Convolutional Block Attention Module (CBAM) [32], among others. Among these, the Attention Gate mechanism proposed by Oktay et al. [9] enhances segmentation accuracy by effectively suppressing irrelevant regions and precisely focusing on salient structures. This mechanism is embedded in the skip connection path of U-Net, where it performs pixel-level fusion of corresponding features from the encoder and decoder to generate attention weight coefficients, thereby guiding the model to concentrate on critical areas. Compared to other typical attention methods, the Attention Gate mechanism is structurally more suitable for the stacking and fusion of multi-path features, exhibiting stronger capability in coordinating multi-scale features.

In terms of feature transmission, ResNet [33] effectively alleviates the gradient vanishing problem in deep networks by introducing residual connections; dense connections [19] further deepen this approach by directly linking shallow and deep features. This not only facilitates gradient backpropagation and feature reuse but also enriches the hierarchical representation of features while increasing network depth. In the task of OCT fluid segmentation, Liu et al. [34] and Hu et al. [35] replaced the original convolutional layers in the encoder-decoder with dense connections, thereby strengthening the feature transmission process and significantly improving the model's segmentation performance.

Regarding context awareness, lesion regions in medical images often exhibit varying sizes and diverse morphologies [36]. Traditional convolutional structures with a single receptive field struggle to adequately capture multi-scale contextual information. To address this issue, Chen et al. proposed the Atrous Spatial Pyramid Pooling (ASPP) module [10]. This module extracts features in parallel using atrous convolutions with multiple sampling rates, effectively integrating cross-scale information ranging from local details to global semantics. It significantly enhances the model's ability to represent targets of different scales. In subsequent studies, Ndipenoch et al. [11] used ASPP to preprocess initial features before feeding them into the network in the nnUNet-RASPP model, while Hu et al. [35] embedded ASPP into the bottleneck layer to further enhance the model's context-aware capabilities. These works collectively validate the effectiveness and applicability of ASPP in complex medical image segmentation tasks.

Therefore, this study integrates an attention gate mechanism, a dense connection structure for enhanced feature transmission, and an Atrous Spatial Pyramid Pooling module to construct an enhanced segmentation network that synergistically combines multi-scale contextual information with multi-level feature representation. In this design, the dense connection structure ensures full reuse and efficient transmission of features across different levels, while the attention gate mechanism enhances the model's focus on key fluid regions through dynamic weight adjustment. This combination enables the network to accurately localize fluid areas while effectively capturing their diverse morphological features, thereby significantly improving segmentation accuracy and generalization for different types of fluids in complex OCT images.

## 2.2. Application of Generative Neural Networks in Medical Imaging

Generative neural networks provide powerful capabilities for data generation and feature extraction in medical image analysis. Among them, Generative Adversarial Networks (GAN) [18] and their variant, Conditional GAN (CGAN) [37], achieve high-quality image synthesis through adversarial training between a generator and a discriminator. The Variational Autoencoder (VAE) [16] learns the probability distribution of data based on an encoder-decoder architecture. Introspective Variational Autoencoder (IntroVAE) [17] innovatively introduces an adversarial mechanism into the VAE framework, where the encoder discriminates the distribution difference of latent variables between real and generated samples, thereby improving reconstruction quality.

Relying on the core mechanism of "reconstructing normality to contrast abnormality," generative models have demonstrated broad potential in medical image segmentation. Such methods typically first use a generative model to reconstruct a corresponding normal tissue image from a pathological image. The lesion area is then accurately delineated by systematically comparing the differences between the original and the generated image. Kobayashi et al. [13], Astaraki et al. [14], and Liu et al. [15] successfully applied GAN, VAE, and IntroVAE respectively, achieving significant results in medical image segmentation tasks such as brain tumors and lung tumors. Furthermore, Shon et al. [38] developed a VAE model specifically for Anterior Segment Optical Coherence Tomography (AS-OCT) images, focusing on analyzing and interpreting the latent anatomical structures within the images. This provides a new perspective for studying anterior segment morphology and highlights the important value of generative models in feature discovery and interpretability analysis.

However, in OCT image analysis, generative neural networks have not yet been applied to the task of fluid region segmentation. Concurrently, IntroVAE maintains the training stability of VAE while significantly improving image generation quality through its built-in adversarial mechanism, effectively overcoming the limitation of blurred outputs in traditional VAEs. Therefore, this study employs IntroVAE to generate high-quality normal OCT images as external prior information. By injecting features from the generated images into the segmentation network, the model is guided to more accurately identify pathological regions, thereby enhancing the precision and robustness of fluid segmentation.

## 3. Method

The proposed Prior-AttUNet model in this study, as illustrated in Figure 2, is designed to specifically address three core challenges in OCT fluid segmentation: inadequate feature representation, variable morphological scales, and low contrast between fluid and normal tissues. To this end, we introduce three key enhancements based on the classical U-Net framework, with the design motivations outlined below:

First, to tackle the issues of inefficient feature transmission and loss of detailed information in deep convolutional networks, we replace the conventional convolutional layers in the encoder with Depthwise Separable Dense Blocks (DenseDepthSepBlock). This design promotes efficient reuse and fusion of features across different levels through dense connections, thereby enhancing gradient flow and preserving richer spatial details, which provides a more powerful feature representation for subsequent accurate segmentation.

Second, to overcome the challenge of multi-scale modeling caused by the diverse shapes and sizes of fluid regions, we incorporate an Atrous Spatial Pyramid Pooling (ASPP) module into the bottleneck layer. By employing parallel atrous convolutions with multiple sampling rates, this module captures contextual information across different receptive fields without down-sampling, enabling the model to simultaneously attend to local details and global semantics. This effectively improves the recognition of fluid regions with varying sizes and shapes.

Finally, and most critically, to address the low contrast and blurred boundaries between fluid and surrounding tissues, we design a Triple Attention Gate mechanism to replace the original skip connections. Inspired by the idea of "reconstructing normality to contrast abnormality," this mechanism dynamically fuses and contrasts high-quality normal OCT images generated by IntroVAE—serving as external prior information—with the detail features from the encoder and the semantic features from the decoder. Through this design, the model actively focuses on the distinctive features between pathological regions and normal anatomical structures, thereby significantly enhancing the identification and segmentation accuracy of critical fluid areas.

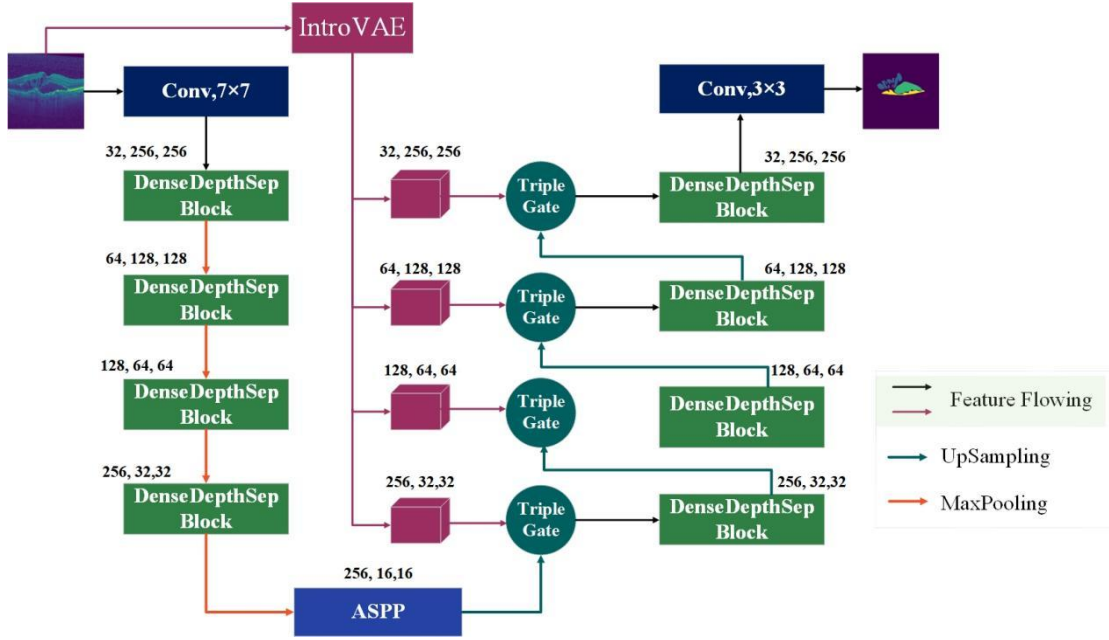

**Figure 2:** Architecture of the Prior-AttUNet. The model enhances a UNet backbone by integrating DenseDepthSepBlock, ASPP, and a triple attention gate, and leverages normal OCT images generated by IntroVAE as prior information to improve focus on fluid regions.

The specific forward propagation process of the model is summarized in algorithmic form as shown in Algorithm 1, which formally outlines the workflow and data interactions of the aforementioned components.

---
**Algorithm 1.**   Prior-AttUNet Forward Propagation
---
**Input:** OCT image $I_{input} \in R_{H \times W \times 3}$
**Output:** Segmentation map $I_{cusput} \in R_{H \times W \times 1}$
**External Prior:** Normal OCT image $I_{prior} \in R_{H \times W \times C}$ generated by IntroVAE
1. //**Encoder Path**
2. $f^0 \leftarrow Conv_{\gamma,T}(I_{input})$ // Extract shallow features
3.   for *l*=1 to 4 do
4.      $f^l_{enc} \leftarrow DenseDepthSepBlock(f^{l-1}_{enc})$ // Feature extraction
5.      $f^l \leftarrow \text{MaxPool}_{2 \times 2}(f^l)$ // Downsampling
6.   end for
7. **Bottleneck**
8. $f_{bottom} \leftarrow \text{ASPP}(f^\epsilon)$ // Multi-scale context extraction
9. // **Decoder Path**

10. $f^4 \leftarrow f_{bottom}$
11. for *l*=4 downto 1 do
12.     $f_{dec}^{l-1} \leftarrow UpSample_{2\times 2}(r_{dec}^l)$   // Upsampling
13.     $f_{an}^{l-1} \leftarrow \text{TripleAttentionGate}(f_{enc}^{l-1}, f_{dec}^{l-1}, l_{prior})$// Feature fusion
14.     $f_{dec}^{l-1} \leftarrow Concat(f_{an}^{l-1}, f_{dec}^{l-1})$// Feature concatenation
15.     $f^{l-1} \leftarrow DenseDepthSepBlock(f^{l-1})$// Feature refinement
16. end for
17. **//Output Layer**
18. $I_{output} \leftarrow Sigmoid(Conv_{2\times 3}(\mathcal{J}_{dec}^q))$// Generate segmentation map
19. return $I_{output}$

The algorithm's main innovation lies in the Triple Attention Gate (step 13), which dynamically integrates encoder details, decoder semantics, and external prior information from IntroVAE. Guided by the "reconstruct normal, compare abnormal" principle, this mechanism enhances focus on pathological regions by contrasting them with normal anatomy. Combined with DenseDepthSepBlock's feature reuse efficiency and ASPP's multi-scale context capture, the architecture forms a robust segmentation model that effectively addresses the challenges of fluid region identification in OCT images.

### 3.1 DenseDepthSepBlock

To achieve efficient feature transmission and reuse in deep networks while optimizing the parameter efficiency of the model, this study employs a Depthwise Separable Dense Block (DenseDepthSepBlock). This module combines the lightweight characteristics of depthwise separable convolution with the high feature reuse capability of a densely connected structure. Its schematic diagram is shown in Figure 3.

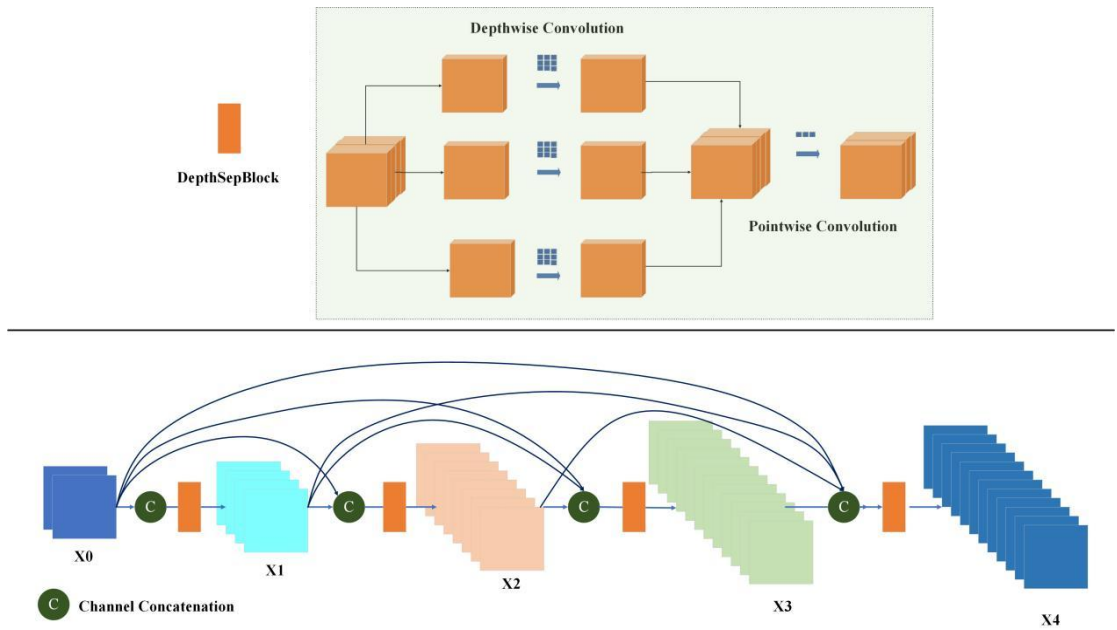

**Figure 3:** DenseDepthSepBlock. This figure is divided into two parts: the upper part illustrates the depthwise separable module, and the lower part depicts the dense connection structure.

Let the input feature map to the module be denoted as $X_0 \in R_{H,W,C_n}$, and the output feature map as $X_L \in R_{H \times W \times C_{out}}$. where $L$ represents the number of convolutional layers within the module (i.e., the ratio). The computational process of the module is as follows:

(1) Depthwise Separable Convolution Layer: Each convolutional layer consists of a Depthwise Convolution followed by a Pointwise Convolution, succeeded by Batch Normalization and a ReLU activation function. The depthwise convolution employs a
3×3 kernel, while the pointwise convolution uses a 1×1 kernel.
(2) Dense Connection: For the $l$-th layer ($L=0,1,…,L−1$), its input is the concatenation of the output feature maps from all preceding layers, i.e.:

$$Z_i = Concat(X_0, X_1, \cdots, X_L) \qquad (1)$$

Where X_0 represents the module input and $X_i$ denotes the output of the i-th layer. The $l$-th layer performs a depthwise separable convolution operation, yielding the output:

$$X_L = ReLU(BN(PointwiseConv(DepthwiseConv(Z_L)))) \qquad (2)$$

By organically integrating depthwise separable convolutions with dense connections, this module significantly reduces the parameter count while promoting cross-layer feature reuse and gradient flow, thereby constructing a rich hierarchical feature representation. Serving as the core building block of the network, the module provides efficient and powerful feature extraction capabilities for the entire segmentation architecture.

### 3.2. ASPP

In the U-Net-based OCT fluid segmentation network, the bottleneck layer serves as a critical component connecting the encoder and decoder, bearing the important responsibilities of feature compression and contextual information extraction. To address the limitations of traditional U-Net bottlenecks, such as their limited receptive field and difficulty in effectively capturing multi-scale fluid features, this study embeds an Atrous Spatial Pyramid Pooling (ASPP) module into the bottleneck layer, as illustrated in Figure 4.

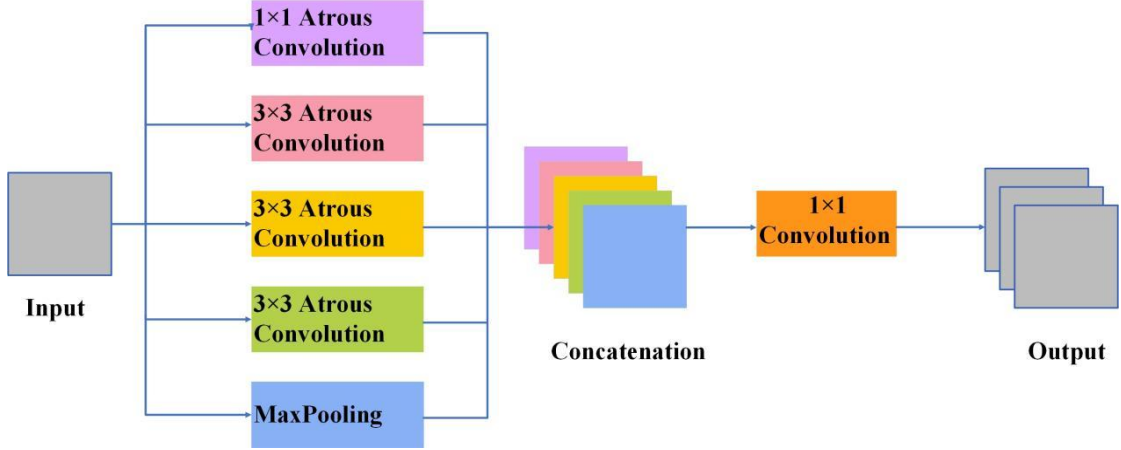

Figure 4. Structure diagram of ASPP. By employing different dilation rates, the module captures local image features at four scales. Combined with the global image information obtained through MaxPooling, it effectively balances both local and global features.

(1) Multi-Scale Reconstruction of Encoder Features.

The encoder, after completing four downsampling operations, produces the feature map {f_{enc}}. At this stage, the spatial dimensions of the input feature map are significantly reduced while the number of channels increases. Introducing the ASPP module at this point offers dual advantages in terms of computational efficiency and feature coverage. Four parallel atrous convolution branches process the high-level semantic features output by the encoder as follows:

$$F_i = Re\,LU(BN(Conv(f_{\varepsilon^i}))), i \in \{1,2,3,4\} \quad (3)$$

Atrous rates are configured as 1, 6, 12, and 18. This expanded sequence of rates is specifically tailored to the reduced spatial dimensions of the bottleneck feature maps, ensuring sufficient receptive field coverage within a limited spatial range.

(2) Global Context Enhancement.

Image-level semantic information is extracted through a global max-pooling branch:

$$F_i = Re\,LU(BN(GMP(\mathcal{J}_{\text{enc}}))) \quad (4)$$

where *GMP* denotes the global max-pooling operation. The feature map output by this branch is upsampled via bilinear interpolation to restore the same spatial dimensions as those of the atrous-convolution branches, thereby providing crucial global prior information for the subsequent decoding process.

(3) Multi-Scale Feature Fusion and Transmission.

The outputs from all branches are concatenated along the channel dimension, followed by feature compression via a 1×1 convolution:

$$F_{concent} = [F_1, F_2, F_3, F_4, F_5] \quad (5)$$

$$F_{bottom} = \text{Dropout}(Re\,LU(BN(Conv(F_{concat})))) \quad (6)$$

The fused features retain rich multi-scale information while maintaining moderate computational complexity, thus providing a solid foundation for the subsequent feature up-sampling and precise localization in the decoder.

### 3.3. IntroVAE

In this study, we employ IntroVAE to reconstruct normal OCT images, thereby generating high-quality normal retinal prior information for the OCT fluid segmentation task. IntroVAE is built upon the variational autoencoder framework and consists of two core components: an encoder and a decoder, as illustrated in Figure 5.

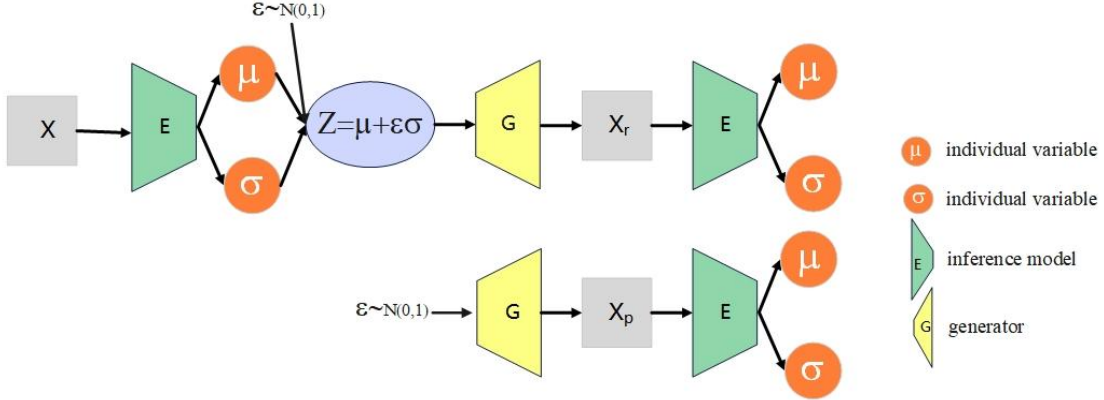

**Figure 5:** The structure of IntroVAE, which consists of two core components: an encoder and a decoder. The encoder is used to map input data into a latent space, learning an implicit representation of the data; the decoder is then used to reconstruct the data from this latent representation.

The encoder maps the input image to the distribution parameters of the latent space:

$$\mu, \log \sigma^2 = Encoder(X) \tag{7}$$

The latent variable is sampled via the reparameterization trick:

$$z = \mu + \epsilon \cdot \exp(0.5 \times \log \sigma^2), \epsilon \sim N(0,1) \tag{8}$$

The decoder reconstructs normal OCT images from the latent variables:

$$X_{rec} = Decoder(z) \tag{9}$$

To progressively capture multi-level feature representations from local details to global semantics, we process the reconstructed normal image through the encoder path of a feature extraction network. This path adopts a four-stage down-sampling structure, where each stage uses a 3 × 3 convolution with a stride of 2 to halve the spatial dimensions while doubling the number of feature channels.

$$f_k^{enc} = Com_{23}^t(f_1^{enc}), k = 1,2,3,4 \tag{10}$$

Among them, Con{v_{3 × 3}^'} denotes the convolutional block at the *l*-th level of the encoder. The output feature map undergoes halving of spatial dimensions and doubling of channel numbers progressively at each stage.

The decoder path performs feature up-sampling via transposed convolutions and incorporates skip connections to fuse features from corresponding encoder levels. At each decoding stage, deeper features are first up-sampled to match the spatial dimensions of the corresponding encoder features. Subsequently, feature fusion is achieved through channel-wise concatenation followed by a 3×3 convolution, thereby progressively restoring spatial detail information:

$$f_1^{dec} = ConvT(Upsample(Concat(f_{i+1}^{dec}, f_{4-l}^{enc}))), l = 3,2,1,0 \tag{11}$$

where Upsample denotes the up-sampling operation, and ConvT represents the transposed convolution.

This encoder-decoder structured feature extraction network further extracts multi-level feature representations from the reconstructed images, providing the subsequent segmentation network with rich information about normal anatomical structures.

During the model training phase, we select normal OCT images without fluid regions from the RETOUCH dataset as training data to ensure the model learns the anatomical features of healthy retinas. Using these normal images, we train IntroVAE in an end-to-end manner by optimizing the reconstruction loss and the regularization term of the latent distribution, enabling the model to accurately capture the morphological patterns of normal retinas. Throughout training, we save the best model weights based on reconstruction performance on the validation set, and load these weights for subsequent reconstruction tasks to generate high-quality normal OCT images.

### 3.4. Triple Attention Gate

To fully leverage the supervision from normal images reconstructed by IntroVAE, this study designs an innovative Triple Attention Gating Mechanism, as illustrated in Figure 6. This mechanism utilizes the regularized features generated by IntroVAE as a key reference to achieve precise localization of fluid regions in OCT images. The design is grounded in the core principle of "reconstructing normality to contrast abnormality," enabling the model to enhance its perception of fluid areas by contrasting pathological regions with normal anatomical structures.

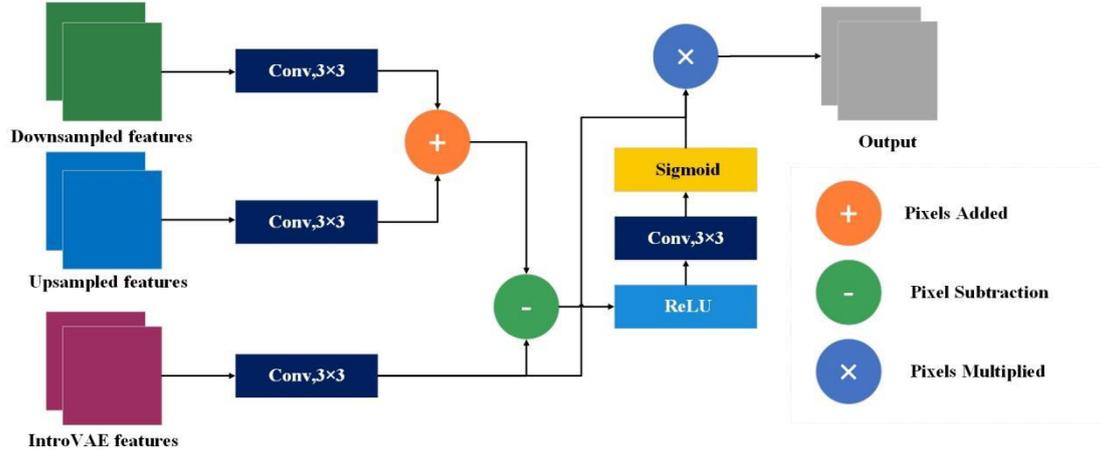

**Figure 6:** Triple Attention Gate. Different from the skip connections in U-Net, the Triple Attention Gate incorporates an additional branch of normal OCT images with prior anatomical features generated by IntroVAE, which enhances the model's focus on effusion regions.

The mechanism first performs feature space alignment on the three input sources. Independent convolutional layers are employed to map the features from different origins into a unified intermediate feature dimension, as specified in Equation (12).

$$X_1 = BN(Conv(f_{\varepsilon e}^l)), X_2 = BN(Conv(f_{dec}^l)). \qquad (12)$$

Here, $f'_{enc}$, $f_{dec}^l$ and $I_{prior}$ denote the encoder features, decoder features, and prior features from the normal image reconstructed by IntroVAE at the *l*-th layer, respectively.

Subsequently, a distinctive addition-subtraction operation is applied to achieve differential fusion of the three feature sources, thereby accentuating the contrast between abnormal regions and normal tissues. This design fully leverages the normal images reconstructed by IntroVAE as a supervisory signal: by subtracting the regularized features from the encoder features, it amplifies the representation of discrepancies between pathological areas and normal anatomical structures. The fusion formula is given by Equation (13).

$$\psi = ReLU(X_1 + X_2 + X_3) \qquad (13)$$

The fused features are then passed through an activation function to generate a spatial attention map, which is used to dynamically re-weight the decoder features. The final output modulates the decoder features through the attention weights, as specified in Equation (14):

$$f_{ann}^l = X_2 \odot \text{Sigmoid}(\Psi) \qquad (14)$$

where $\odot$ denotes element-wise multiplication.

This triple attention gate mechanism effectively enhances the model's perception of fluid regions by fusing multi-source information and introducing a differential contrast mechanism. Compared to traditional dual-source attention mechanisms, the VAE-generated features incorporated here provide important normal anatomical priors, enabling the model to more accurately distinguish between pathological regions and healthy tissues.

## 4. Experiment and Result
### 4.1. Dataset

The experiments in this study were conducted based on the dataset from the MICCAI 2017 RETOUCH challenge [7]. This dataset contains images acquired from three different OCT imaging devices (Cirrus, Spectralis, and Topcon), which exhibit significant variations in key parameters such as the number of scanning channels and image resolution, as detailed in Table 1. This multi-source heterogeneous nature poses a challenge to the generalization capability of models. Through slice-wise processing of the samples, a total of 3306 annotated slice images were obtained. The experiments employed a 4:1 ratio to split the data into a training set and a test set.

**Table 1:** Dataset Statistics and Class Distribution

| Device | Subjects | Device | Slices | Background (%) | IRF(%) | SRF(%) | PED(%) |
|---|---|---|---|---|---|---|---|
| Cirrus | 24 | 512×1024 | 3072 | 98.95 | 0.04 | 0.18 | 0.83 |
| Spectralis | 24 | 512×496 | 1176 | 98.74 | 0.15 | 0.39 | 0.72 |
| Topcon | 22 | 512×885 | 2688 | 99.45 | 0.03 | 0.06 | 0.46 |

As clearly observed from Table 1, there exists a severe class imbalance in the data: background regions dominate overwhelmingly across all devices (98.74%–99.45%), while the combined proportion of the three lesion regions (IRF, SRF, PED) is consistently less than 1.5%. It is particularly noteworthy that the proportion of IRF is below 0.2% in all devices, the highest proportion of SRF is 0.39% (Spectralis), and PED, though relatively higher, does not exceed 0.83%. The distribution of lesions also varies across devices: the total proportion of fluid regions is relatively higher in Spectralis, whereas Topcon shows the lowest proportion for each lesion type. Such extreme class imbalance and cross-device variability pose significant challenges for model training.

### 4.2 Evaluation Metrics

This study employs commonly used evaluation metrics in the field of image segmentation for quantitative analysis, including the Dice Similarity Coefficient (*DSC*) and the mean Dice Similarity Coefficient (*mDSC*). The DSC is used to evaluate the overlap between the segmentation results of a specific category (background, IRF, SRF, or PED) and the ground truth annotations, while the *mDSC* represents the macro-average of the segmentation overlap across all categories. The specific calculation formulas are as follows.

The formula for *DSC* is shown in Equation (15):

$$DSC = (2^*|X \cap Y|) / (|X| + |Y|) \qquad (15)$$

Where *X* and *Y* represent the model-predicted segmentation region and the ground-truth annotation region, respectively, and $|\cdot|$ denotes the number of pixels in the set.

The formula for *mDSC* is given in Equation (16):

$$mDSC = (1/N) * \sum_{i=1}^{N} DSC_{-i} \qquad (16)$$

where *N* is the total number of categories, and *DSC_i* is the Dice coefficient for the *i*-th category.

Both *DSC* and *mDSC* values range from 0 to 1. A higher value indicates greater consistency between the model's predicted segmentation and the ground-truth annotations, representing better segmentation performance.

### 4.3. Implementation

This study was conducted on a workstation equipped with an Intel Core i9-14900K processor and an NVIDIA RTX 4090 GPU (24 GB VRAM). The software environment was built on the PyTorch framework, with the integrated development environment being PyCharm 2024.1.4 (Community Edition). Key software versions included Python 3.12, PyTorch 2.7.0+cu126, TorchVision 0.22.0+cu126, and NumPy 1.26.4.

During the data preprocessing stage, input images were resized to 256 × 256 pixels using nearest-neighbor interpolation and normalized to the range [0, 1]. The dataset was randomly split into training and test sets at an 80:20 ratio. During training, data were loaded with a batch size of 16 and randomly shuffled. The AdamW optimizer was employed with an initial learning rate of 0.0001, momentum parameters β set to (0.9, 0.99), and a numerical stability term ε of $1 \times 10^{-8}$; the batch size during training was fixed at 16.

To address the class-imbalance problem, the loss function combined Lovasz Loss and Dice Loss. Dice Loss directly measures the region overlap between the predicted segmentation and the ground-truth labels, reducing the model's reliance on the number of foreground pixels and thus effectively mitigating the imbalance caused by the overwhelming majority of background pixels compared to target regions such as IRF and SRF. Meanwhile, Lovasz Loss, as a differentiable convex relaxation upper bound of the Dice coefficient, provides smoother and more robust gradient information during back-propagation. The two losses work synergistically in the optimization process, not only enhancing the stability of model training but also guiding the model toward better convergence in overall segmentation performance, especially in the boundary localization accuracy of minority classes.

Dice Loss is based on the Dice similarity coefficient, as shown in Formula (17).

$$L_{\text{DCE}} = 1 - \frac{2 \sum_i p_i g_i + \epsilon}{\sum_i p_i + \sum_i g_i + \epsilon} \qquad (17)$$

where $p_i$ is the predicted probability that the *i*-th pixel belongs to the target class, $g_i$ is the corresponding ground-truth label, and $\varepsilon$ is a smoothing term used to prevent division by zero errors.

Lovasz Loss aims to directly optimize a surrogate loss for the mean Intersection over Union (mIoU). It leverages the Lovász extension to transform the non-differentiable mIoU into a differentiable convex function for optimization. For multi-class segmentation problems, the Lovasz Loss computes the Jaccard index loss induced by prediction errors for each class and sums them, as shown in Equation (18).

$$L_{\text{Lovaix}} = \frac{1}{|C|}\sum_{c \in C} \mathcal{A}_{J_c}(m(c)) \tag{18}$$

Where $|C|$ denotes the total number of classes, and $\mathcal{A}_{J_c}$ represents the convex Lovász extension of the Jaccard loss for class c computed based on the sorted components of the error vector m(c). This formulation enables direct and continuous optimization of the IoU metric within a gradient-based learning framework, improving both boundary-aware segmentation quality and training stability.

Ultimately, the model underwent 150 training epochs, with the peak mean Dice Similarity Coefficient (*mDSC*) achieved on the test set serving as the performance evaluation criterion. The corresponding best-performing weights were preserved for subsequent analysis.

## 4.4. Results

In the field of OCT fluid segmentation, our proposed Prior-AttUNet method demonstrates outstanding performance. We conducted comparative experiments with current state-of-the-art methods across three imaging devices: Cirrus, Spectralis, and Topcon. The results indicate that Prior-AttUNet achieves a significant breakthrough in segmentation accuracy.

As shown in Table 2, Prior-AttUNet achieved the best results in terms of the mDSC metric across all three devices, specifically: Cirrus (93.93 ± 1.6 %), Spectralis (95.18 ± 0.3 %), and Topcon (93.47 ± 0.3 %). Notably, on the Spectralis and Topcon devices, its remarkably low uncertainty (±0.3 %) demonstrates excellent stability. This performance clearly surpasses that of the recent top-performing method, DAA-UNet (by an average margin of approximately 3 percentage points), and far exceeds the baseline methods such as the conventional UNet and its variants.

Regarding computational efficiency, Prior-AttUNet also performs excellently. Its computational complexity is only 0.37 TFLOPs, which, while slightly higher than Deeplabv3+, is considerably lower than other high-performance methods (e.g., MsTGANet's 1.12 TFLOPs and DAA-UNet's 1.28 TFLOPs), demonstrating superior inference efficiency. Although its parameter count of 47.04M is higher than some lightweight designs, it remains within an acceptable range. Careful architectural optimization ensures a well-balanced trade-off between model performance and complexity.

Overall, Prior-AttUNet achieves the best balance between accuracy and efficiency for the OCT fluid segmentation task, providing a reliable solution for clinical applications that demand both high accuracy and real-time performance.

It should be noted that, as RetiFluidNet, FAM-U-Net, and FNeXter have not made their model code publicly available, we were unable to calculate their FLOPs and parameter counts. Furthermore, RetiFluidNet and FAM-U-Net report a weighted average of the DSC metrics for the three fluid types: IRF, SRF, and PED.

TABLE 2: Comparison of *mDSC* for different methods on three imaging devices

| Method | Year | *mDSC*(%) | | | FLOPs (TFLOPs) | Params(M) |
|---|---|---|---|---|---|---|
| | | Cirrus | Spectralis | Topcon | | |
| UNet [8] | 2015 | 75.7±4.4 | 75.8±5.1 | 77.1±5.5 | 0.41 | 21.63 |
| Attention-UNet [9] | 2018 | 79.6±3.9 | 74.7±4.6 | 81.0±4.9 | 0.44 | 22.33 |
| Deeplabv3+ [10] | 2018 | 83.6±2.4 | 79.2±3.9 | 81.7±3.5 | **0.33** | 54.51 |
| MsTGANet [12] | 2021 | 86.6±1.5 | 86.3±1.4 | 85.7±2.9 | 1.12 | **11.60** |
| RetiFluidNet [5] | 2023 | 87.2±2.1 | 90.5±1.3 | 85.1±3.4 | - | - |
| nnUnet RASPP [11] | 2023 | 84.1±2.7 | 85.2±2.4 | 82.7±3.3 | 0.91 | 22.11 |
| FNeXter [39] | 2024 | 82.3±0.46 | 83.1±0.93 | 77.5±0.6 | - | - |
| FAM-U-Net[40] | 2024 | 87.7 ± 3.1 | 85.1±3.7 | 85.8±2.7 | - | - |
| DAA-UNet[35] | 2025 | 90.2±1.4 | 91.6±1.0 | 90.5±1.2 | 1.28 | 17.44 |
| RSAPower[6] | 2025 | 81.98±1.1 | 83.56±3.0 | 86.38±4.7 | 0.96 | 18.41 |
| **Prior-AttUNet(ours)** | - | **93.93±1.6** | **95.18±0.3** | **93.47±0.3** | 0.37 | 47.04 |

As shown in Figure 7, the comparative segmentation results visually demonstrate the outstanding performance of Prior-AttUNet. Overall, the model's predictions exhibit high similarity to the ground-truth annotations, corroborating the leading performance achieved in the quantitative evaluation. Specifically, in cases from all three devices, Prior-AttUNet is able to completely and accurately segment fluid regions of varying morphologies (typically displayed as bright white areas in the figure). The predicted boundaries are clear and continuous, almost overlapping with the contours of the ground-truth labels. Particularly in structurally complex regions, the model also shows excellent recognition capability, with no obvious over-segmentation or under-segmentation.

Benefiting from the fusion of normal anatomical priors and pathological features enabled by the triple attention gate mechanism, the model can precisely perceive and enhance boundary information of fluid regions. Meanwhile, the multi-scale architecture ensures robust segmentation performance across different imaging characteristics of the devices. The cross-device consistent high-similarity segmentation effects shown in the figure strongly reinforce the quantitative conclusions reported earlier, such as "achieving the best results on all three devices" and "extremely low uncertainty," thereby fully validating the effectiveness and reliability of the Prior-AttUNet method in practical applications.

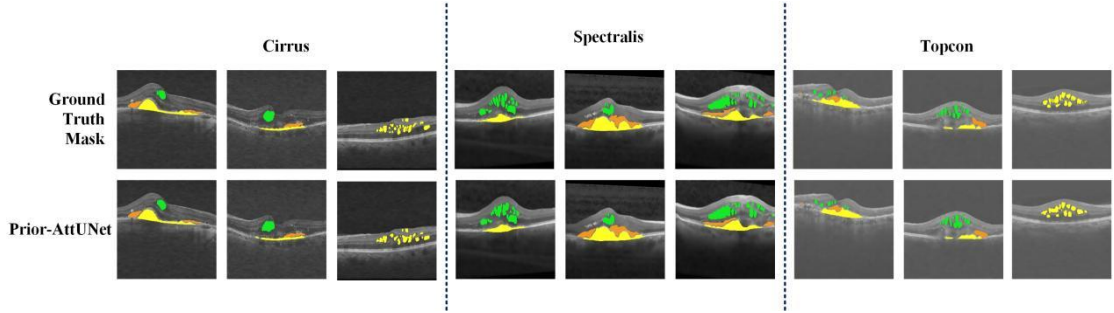

**Figure 7:** Visualization of segmentation results. The first row shows the ground-truth masks, and the second row below displays the corresponding model predictions. From left to right, each image group originates from the three different imaging devices: Cirrus, Spectralis, and Topcon.

### 4.5 Ablation Experiments

To comprehensively evaluate the effectiveness of the Prior-AttUNet model, we conducted a systematic ablation study. All experiments were performed on a dataset acquired using the Spectralis device, which contains 1176 retinal OCT images. We randomly split the training set at a ratio of 80%. All models were trained for 150 epochs, and the model weights achieving the highest average Dice coefficient (AVG) were retained for evaluation on the test set. Performance is reported using four metrics: SRF (%), IRF (%), PED (%), and AVG (%), ensuring comparability of the experimental results.

#### 4.5.1. Ablation Studies on Components

To verify the effectiveness of each core component in the proposed Prior-AttUNet model, we designed systematic ablation experiments. Using our complete proposed model as the baseline, performance comparisons were conducted by sequentially removing key modules. The detailed results are presented in Table 3.

**Table 3:** Ablation Study of Architectural Components in Prior-AttUNet

| Model Variant | SRF(%) | IRF(%) | PED(%) | AVG(%) |
|---|---|---|---|---|
| w/o NORMNET(No Anatomical Prior) | 0.9151 | 0.9635 | 0.8526 | 0.9325 |
| w/o Dense Blocks (Standard Convs) | 0.9256 | 0.9582 | 0.8674 | 0.9375 |
| w/o Triple Attention | 0.9404 | 0.9651 | 0.8766 | 0.9452 |
| w/o ASPP Module | 0.9370 | 0.9612 | 0.8835 | 0.9451 |
| **Prior-AttUNet** | **0.9607** | **0.9729** | **0.8927** | **0.9563** |

First, the normal anatomical prior plays a crucial role in model performance. Removing this module (w/o IntroVAE) resulted in a decrease of 2.38% in the average Dice coefficient, with the most significant performance drop observed for PED segmentation (a reduction of 4.01%). This indicates that the prior, serving as a reference baseline, effectively enhances the model's ability to represent features of lesion regions through contrastive learning, which is vital for identifying complex boundaries. Second, the triple attention mechanism also demonstrates a significant contribution. After removing this module (w/o Triple Attention), the Dice coefficients for SRF and PED decreased by 2.03% and 1.61%, respectively. This validates its ability to achieve adaptive feature selection by fusing multi-source features, thereby effectively improving boundary segmentation accuracy.

Regarding multi-scale context and feature reuse, both the ASPP module and the dense connection blocks provide important auxiliary benefits. Removing the ASPP module (w/o ASPP Module) led to reductions of 2.37% and 1.17% for SRF and IRF, respectively, highlighting the importance of multi-scale contextual awareness for recognizing lesion regions of varying sizes. Similarly, replacing the dense connection blocks with standard convolutions (w/o Dense Blocks) also reduced the average Dice coefficient by 1.88%, indicating that this design enhances model robustness by improving feature reuse and propagation efficiency.

The experimental results show that each core component substantively contributes to performance improvement, with the normal anatomical prior and the triple attention mechanism being the most critical. This effectively validates the efficacy of the main innovations proposed in this paper.

### 4.5.2. Ablation Study on the NORMNET Architecture

This ablation study aims to evaluate the specific contribution of the NORMNET architecture design. NORMNET is a normal anatomical prior module generated by IntroVAE, providing four levels of feature representation. The experimental results demonstrate that the module's design, based on the IntroVAE reconstructor, is irreplaceable in delivering high-quality anatomical priors.

As shown in Table 4, all alternative solutions in the module replacement experiments led to performance degradation. When NORMNET was replaced with simple down-sampling, the model's Dice coefficient for PED segmentation dropped to 86.89%, representing a decrease of approximately 2.66% compared to the complete model (89.27%). This indicates that simple feature extraction cannot generate effective normal anatomical priors for contrastive learning. Replacing it with a ResNet feature extractor resulted in a particularly significant drop in the SRF metric—from 96.07% to 93.17%, a reduction of approximately 3.02%—demonstrating that generic feature extractors struggle to learn discriminative patterns of normal anatomy.

In the module capacity experiments, adjusting the channel count of NORMNET consistently impaired performance. Reducing the channels from 32 to 16 lowered the average Dice coefficient from 95.63% to 93.95%, a drop of about 1.68%, suggesting that insufficient channels hinder adequate encoding of prior information. Conversely, increasing the channel count to 64 also reduced the average Dice coefficient to 94.11%, a decline of about 1.52%, indicating that excessive channels may introduce redundancy and increase overfitting risk. These results confirm that the current 32-channel design achieves an optimal balance between model capacity and efficiency.

This experiment validates the rationale behind the NORMNET design from two dimensions: module architecture and capacity. The reconstruction-based architecture of IntroVAE proves superior to both simple down-sampling methods and generic feature-extraction networks in providing anatomical priors, while its specific channel configuration is demonstrated to be the optimal choice. Any simplification, replacement, or inappropriate capacity adjustment of the architecture impairs the quality of prior generation and consequently degrades overall segmentation performance. This strongly supports the originality of our design for this module within the generative-segmentation dual-path architecture.

**Table 4: Ablation Study on the NORMNET Architecture**

| Model Variant | SRF(%) | IRF(%) | PED(%) | AVG(%) |
|---|---|---|---|---|
| Replacing NORMNET with simple downsampling | 0.9524 | 0.9661 | 0.8689 | 0.9465 |
| Replacing NORMNET with a ResNet feature extractor | 0.9317 | 0.9696 | 0.8885 | 0.9472 |
| Reducing NORMNET channels (midc: 32→16) | 0.9368 | 0.9568 | 0.8657 | 0.9395 |
| Increasing NORMNET channels (midc: 32→64) | 0.9327 | 0.9631 | 0.8697 | 0.9411 |
| Prior-AttUNet (Full Model) | **0.9607** | **0.9729** | **0.8927** | **0.9563** |

### 4.5.3. Ablation Study on the Triple Attention Gate

This ablation study aims to validate the effectiveness of the proposed Triple Attention Gate design. By comparing it with various mainstream attention mechanisms and their variants, the results demonstrate that our method holds a significant advantage in segmentation accuracy.

As shown in Table 5, in the comparison with other attention mechanisms, our proposed method demonstrates comprehensive superiority. The concatenation-based attention gating achieves an average Dice coefficient of **93.99%**, which is **1.64%** lower than our method (**95.63%**); the SAM mechanism shows comparable overall performance but exhibits a significant drop of **3.37%** in PED segmentation; the CAM mechanism yields the lowest performance, with an average Dice coefficient of only **89.68%**, representing a decrease of **5.95%**; while the CBAM mechanism performs comparably on IRF, it declines by **4.10%** and **4.13%** on SRF and PED, respectively. These results indicate that existing attention mechanisms struggle to achieve fine-grained feature selection in complex medical image segmentation tasks.

Critically, when we removed the anatomical prior input generated by NORMNET (i.e., using only the decoder feature $g$ and the skip-connection feature $x$), performance was significantly affected, with the average Dice coefficient dropping by 0.0238. This outcome strongly demonstrates the necessity of incorporating normal anatomical priors as the third input source—they provide an indispensable reference baseline for core segmentation decisions.

**Table 5:** Ablation Study on Attention Mechanisms

| Model Variant | SRF(%) | IRF(%) | PED(%) | AVG(%) |
|---|---|---|---|---|
| Concatenation-based Attention | 0.9306 | 0.9633 | 0.8671 | 0.9399 |
| SAM | 0.9337 | 0.9572 | 0.8590 | 0.9371 |
| CAM | 0.8299 | 0.9437 | 0.8156 | 0.8968 |
| CBAM[27] | 0.9197 | 0.9636 | 0.8514 | 0.9333 |
| w/o NORMNET Input (Only g & x) | 0.9151 | 0.9635 | 0.8526 | 0.9325 |
| Triple Attention Gate (Ours) | 0.9607 | 0.9729 | 0.8927 | 0.9563 |

### 4.5.4. Ablation Study on Dense Connectivity

This ablation study systematically evaluates the impact of design choices for the Dense Blocks on both model performance and efficiency. The results demonstrate that our final design—a lightweight dense block based on depthwise separable convolutions with a ratio=3—achieves the optimal balance between performance and efficiency.

As shown in Table 6, regarding the choice of convolution types, replacing depthwise separable convolutions with standard convolutions significantly increased the model's computational burden—FLOPs rose from 0.37 to 0.59 TFLOPs, and the number of parameters increased from 47.04M to 51.70M, yet the average segmentation performance decreased by 1.71%. This demonstrates that depthwise separable convolutions substantially enhance the model's computational efficiency with almost no compromise in performance, validating their core value in our lightweight design.

In contrast, completely replacing the dense connection blocks with residual blocks (ratio=1) resulted in an average performance drop of 0.0086. Although it had the lowest computational cost (0.34 TFLOPs), the performance sacrifice confirms that dense connections hold an irreplaceable advantage in promoting feature reuse and propagation.

A further analysis of connection intensity clearly indicates the existence of a performance-optimal "sweet spot." When connections are insufficient (ratio=2), the average performance decreases by 1.18%, suggesting inadequate feature reuse. Conversely, when connections are excessive (ratio=4 & 5), performance significantly declines (by 2.03% and 2.56%, respectively), while computational costs continue to rise. This shows that overly dense connections may complicate network optimization and introduce feature redundancy, ultimately impairing model performance.

**Table 6:** Ablation Study on the Design of Dense Connections

| Model Variant | SRF(%) | IRF(%) | PED(%) | AVG(%) | FLOPs (TFLOPs) | Params (M) |
|---|---|---|---|---|---|---|
| Using standard convolution in dense blocks | 0.9219 | 0.9634 | 0.8727 | 0.9392 | 0.59 | 51.70 |
| Prior-AttUNet (ratio=1) | 0.9411 | 0.9701 | 0.8807 | 0.9477 | 0.34 | 46.51 |
| Prior-AttUNet (ratio=2) | 0.9441 | 0.9605 | 0.8748 | 0.9445 | 0.35 | 46.72 |
| Prior-AttUNet (ratio=3) | **0.9607** | **0.9729** | **0.8927** | **0.9563** | **0.37** | **47.04** |
| Prior-AttUNet (ratio=4) | 0.9267 | 0.9550 | 0.8634 | 0.9360 | 0.39 | 47.78 |
| Prior-AttUNet (ratio=5) | 0.9248 | 0.9453 | 0.8540 | 0.9307 | 0.42 | 48.04 |

### 4.5.5. Ablation Study on Multi-scale Context Aggregation

This ablation study evaluates the role of the ASPP module in extracting multi-scale contextual information. The experimental results indicate that ASPP plays a crucial role in improving model performance, particularly in segmenting SRF regions with diverse shapes.

As shown in Table 7, when replacing ASPP with the Pyramid Pooling Module (PPM) from PSPNet, the model's average performance declined, with the Dice coefficient dropping from 95.63% to 94.20%, a decrease of 1.43%. Among these, the performance in SRF segmentation showed the most significant decline, falling from 96.07% to 92.76%, a reduction of 3.31%. This discrepancy indicates that although PPM also possesses multi-scale perception capabilities, ASPP—by expanding the receptive field through dilated convolutions while maintaining feature map resolution—is more suitable for retinal fluid segmentation tasks that require precise boundary localization.

Furthermore, when the multi-scale module was completely removed and only ordinary convolutional layers were used, the model performance further dropped to 94.03%—a decline of 1.60% compared to the complete model (95.63%). This confirms the universal importance of multi-scale context aggregation for fundus OCT image segmentation—the varying sizes of fluid regions necessitate the model's ability to simultaneously perceive local details and understand global context.

A comprehensive analysis shows that the ASPP module, with its unique atrous convolution design, effectively aggregates multi-scale feature information while preserving spatial accuracy.

**Table 7: Ablation Study on Multi-scale Context Modules**

| Model Variant | SRF(%) | IRF(%) | PED(%) | AVG(%) |
|---|---|---|---|---|
| Replacing ASPP with PPM[41] | 0.9276 | 0.9638 | 0.8777 | 0.9420 |
| w/o ASPP (Standard Convolution) | 0.9276 | 0.9592 | 0.8757 | 0.9403 |
| Prior-AttUNet with ASPP | 0.9607 | 0.9729 | 0.8927 | 0.9563 |

## 4.6. Visualization Analysis of Attention in the Decoder Path

The heatmap sequence shown in Figure 8 visually reveals the progressive evolution and focusing of attention during the decoding process of the Prior-AttUNet model.

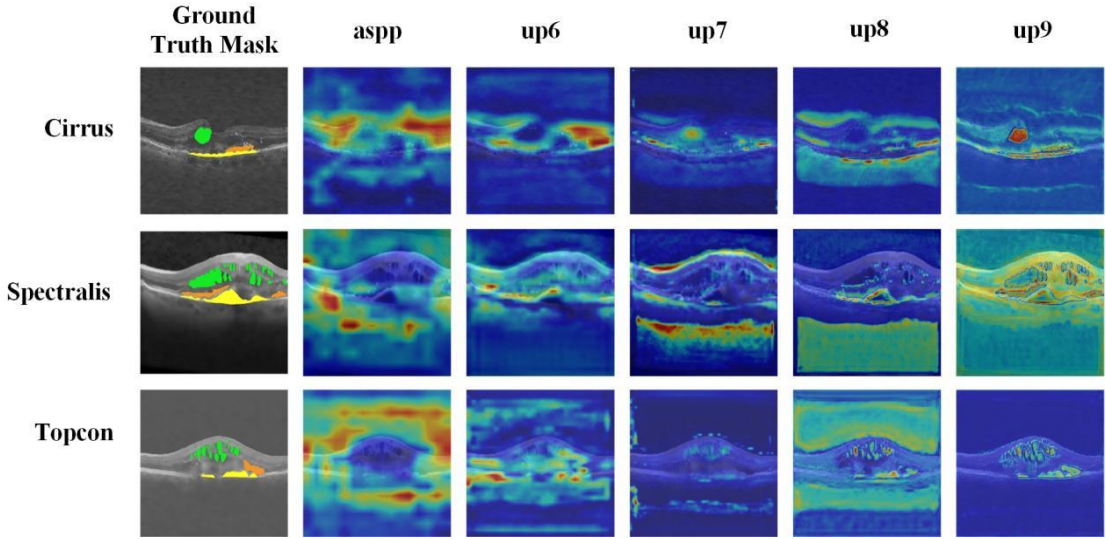

**Figure 8:** Visualization of attention along the decoder path. Each row represents a case from a different device, arranged from top to bottom as Cirrus, Spectralis, and Topcon. From left to right, each row displays: the ground truth mask, along with the feature heatmaps generated by the model at the bottleneck ASPP module and the subsequent up-sampling stages (up6, up7, up8, up9).

As clearly illustrated by the evolving trend of the heatmaps, the model's attention regions undergo a progressive refinement from broad to precise focus as decoding depth increases. Specifically, in the earlier ASPP and up6 stages, the heatmaps display relatively dispersed activation areas, indicating that the model is conducting extensive perception and screening of global context and potential structures. As up-sampling proceeds (through the up7 and up8 stages), the highlighted response areas rapidly converge and closely align with the morphology of the true fluid regions. By the final up9 stage, which is closest to the output, the highlighted portions of the heatmap become highly concentrated and accurately cover the target fluid areas, exhibiting remarkable consistency with the contours of the ground-truth annotations. This demonstrates that the model effectively directs its attention toward the pathological regions before making final decisions.

This visualization strongly validates that the triple attention gate mechanism, in synergy with multi-scale feature fusion, guides the model to adaptively suppress irrelevant background interference during the decoding path while progressively enhancing the capture and localization of target pathological features (fluid), ultimately achieving precise segmentation. The consistent attention-focusing patterns observed across cases from different devices further confirm that the model's internal feature selection mechanism maintains stable robustness when confronted with varying imaging device characteristics.

## 5. Discussion and Conclusion

Medical image segmentation faces persistent challenges due to the complexity of imaging mechanisms, anatomical structures, and pathological features. The Prior-AttUNet framework proposed in this study effectively addresses the specific difficulties in fluid segmentation within retinal OCT images by introducing generative anatomical priors and a triple attention mechanism. Comprehensive evaluation on OCT datasets from three different imaging devices—Cirrus, Spectralis, and Topcon—demonstrates that our method achieves mDSC scores of 93.93%, 95.18%, and 93.47%, respectively. These results significantly surpass various state-of-the-art methods, including FAM-U-Net (2024), RSAPower (2025), and DAA-UNet (2025), fully affirming the framework's strong generalization capability and robustness across different OCT devices.

The core advantage of our model stems from its generative-prior-guided synergistic architecture design. Ablation experiments confirm that the anatomical prior module (NORMNET) contributes an average performance improvement of 2.38%, with segmentation enhancement for the complex structure PED reaching 4.01%. The triple attention mechanism provides accuracy gains of 2.03% and 1.61% for SRF and PED boundary segmentation, respectively. Systematic component optimization has established the optimal configuration: the depthwise-separable-convolution-based dense block (ratio=3) achieves efficient feature reuse within a computational budget of 0.37 TFLOPs; the ASPP module outperforms PPM by 3.31% in SRF segmentation, validating the boundary-aware advantage of its multi-scale design; and the 32-channel configuration of NORMNET is proven to strike an optimal balance between feature representation and model complexity.

It is worth noting that, although our model demonstrates outstanding segmentation accuracy, its parameter count (47.04 M) still leaves room for optimization compared to some baseline methods, primarily due to the dense-connection design adopted to preserve feature representation capability. Future work will focus on maintaining model performance while further exploring lightweight technical pathways such as neural architecture search, model pruning, and knowledge distillation, aiming to enhance computational efficiency and better meet the demands of real-time clinical applications.

In summary, the proposed Prior-AttUNet achieves significant performance improvement in retinal OCT image fluid segmentation through the organic integration of generative-prior enhancement and adaptive feature selection. The innovation of this framework lies in systematically introducing the concept of normal-anatomical-prior guidance into the field of OCT fluid segmentation for the first time, providing a new perspective for addressing the challenges of this specific task. The comprehensive ablation study not only verifies the effectiveness of each module but also offers reliable methodological references for the design of similar models. Future research will be dedicated to improving the computational efficiency of the model and exploring its potential for cross-organ generalization in a wider range of medical image segmentation tasks.

**CRediT authorship contribution statement**

**Li Yang**: Writing-review & editing, Conceptualization of this study,Methodology, Software. **Yuting Liu**: Conceptualization of this study, Methodology,Soft-ware, Project administration.

**Acknowledgments**

This work was supported by the the Scientific Research Project of Higher Education Institutions in Anhui Province (2024AH053451, 2024AH053462), the Horizontal Research Project of Wannan Medical College (H202530).